\documentclass[11pt]{article}

\usepackage[preprint]{acl}
\usepackage{times}
\usepackage{latexsym}
\usepackage[T1]{fontenc}
\usepackage[utf8]{inputenc}
\usepackage{microtype}
\usepackage{inconsolata}
\usepackage{graphicx}
\usepackage{amsmath}
\usepackage{amssymb}
\usepackage{booktabs}
\usepackage{float}

\title{SAGE: Governed Artifact Generation from Enterprise Guidelines}
\author{
  Mohammadreza Sediqin \\
  Centific Research \\
  New York, USA \\
  \texttt{\small mohammadreza.s@centific.com} \\\And
  Shivali Dalmia \\
  Centific Research \\
  Seattle, USA \\
  \texttt{\small shivali.dalmia@centific.com} \\\And
  Sumukha Thoppanahalli \\
  Centific Research \\
  Seattle, USA \\
  \texttt{\small sumukhasharma.t@centific.com} \\\AND
  Srinivasa Karthikeya Reddy Kovvuri \\
  Centific Research \\
  Washington, USA \\
  \texttt{\small srinivasa.kovvuri@centific.com} \\\And
  Abhishek Mukherji \\
  Centific Research \\
  Seattle, USA \\
  \texttt{\small abhishek.mukherji@centific.com}
}

\begin{document}

\maketitle
\begin{abstract}
Enterprise guideline documents mix narrative text, complex tables, and embedded images, and converting them into structured work artifacts still takes two to three days of manual effort each. Current language and vision-language models extract from such documents but offer no governed workflow beyond extraction: no validation, no consistency checking, no traceable artifact generation. We introduce SAGE, a governed multi-stage LLM pipeline organized around a shared versioned rule store with stable identifiers, schema-validated inter-stage contracts, and end-to-end provenance tracking. Extracted rules undergo deterministic structural validation and LLM-based semantic scoring, then a consistency module that removes duplicates, flags contradictions, and surfaces specification gaps; only uncertain or flagged items reach reviewers, while high-confidence outputs are auto-approved. On 120 documents, SAGE cuts turnaround from days to 20--100 minutes, achieving a 96\% document-level success rate with 3.2\% hallucination, extracting 3,896 rules and producing 812 artifacts ready for human review; without governance, hallucination rises to 15.7\%.
\end{abstract}

\section{Introduction}
\label{sec:intro}
In modern enterprise pipelines, annotation projects rely on unstructured guideline documents that must be converted into structured, executable work artifacts before any labeling can begin. This conversion is carried out manually by quality managers (QM) and project managers (PM), who read through the guidelines, interpret implicit rules, resolve ambiguous cases, and assemble deliverables such as annotator instructions and statements of work. Each document typically takes two to three days, often yields inconsistencies and errors, and must be redone from scratch whenever the source is revised~\cite{anderson2024design, perot2024lmdx}, delaying staffing, launch, and ongoing maintenance across clients.

Extraction is difficult here because these documents are multimodal and structurally irregular. Text is frequently represented as positional tokens rather than coherent semantic units, complicating layout reconstruction. Tables may extend across pages, include merged cells, or appear purely as images with no underlying structure. Labeling examples, bounding box diagrams, and edge case illustrations often carry critical information that appears nowhere in the document text~\cite{ke2025large, bhattacharyya2025information,dalvand2025regional}. In production, any parsing error cascades directly into downstream rule quality.

Production workflows also impose requirements that current systems do not address. Extracted rules must be validated for structural soundness and semantic quality, filtered for consistency before storage, reconciled across document versions, then converted into persona-specific artifacts with provenance back to their source rules. Parsers such as Docling~\cite{livathinos2025docling} and instruction-tuned models such as UIE~\cite{lu2022unifiedstructuregenerationuniversal} target extraction alone, while LLM and VLM based approaches~\cite{bai2025qwen3, wang2024qwen2} support multimodal reasoning but struggle with complex layouts and remain vulnerable to hallucination. To our knowledge, no existing system combines a shared versioned rule store, schema-validated inter-stage contracts, layered automatic evaluation, and threshold-driven persona-routed human review into a single governed pipeline.

We introduce SAGE, a governed multi-stage LLM pipeline in which multiple LLM calls coordinate through a shared versioned rule store to convert raw enterprise guideline documents into structured, validated operational artifacts. At its core this is a data management problem: each stage reads and writes schema-validated objects keyed by a stable \texttt{rule\_id}, functioning as a typed intermediate relation rather than a free-form message queue. This enforces contractual guarantees between stages and yields full provenance, so every artifact traces back to its source rule and originating document. Because client guidelines fall under non-disclosure agreements, all stages that process raw client documents use self-hostable models (Qwen2.5-VL-32B, Qwen3-32B, LLaVA-13B); later stages operate only on extracted rule representations rather than source content.

Our contributions are as follows:

\begin{itemize}
\item \textbf{SAGE}, a governed multi-stage LLM pipeline spanning deterministic parsing, VLM-based extraction, structured rule modeling, and dependency-aware artifact generation.

\item \textbf{A two-stage evaluation framework} combining deterministic
structural validation (L1) with LLM-based semantic scoring (L2), driving
automated acceptance, targeted regeneration, and selective Human-in-the-Loop (HITL) escalation.

\item \textbf{A dependency-driven HITL workflow} over rules, gaps, and
artifacts, with zero-edit approvals reused as calibration signals for the
evaluation judge, progressively reducing review load over deployment cycles.

\item \textbf{Evaluation on 120 real-world enterprise guideline documents},
where SAGE extracts 3,896 rules and produces 812 artifacts at 3.2\%
hallucination, against 15.7\% for an ungoverned one-pass baseline.
\end{itemize}

\section{Related Work}
\label{sec:related}

\subsection{Document Parsing and Structured Extraction}
Document parsing has evolved from rule-based OCR~\cite{smith2007overview, cui2021document} to hybrid vision-language approaches spanning text, layout, and visual modalities~\cite{poznanski2025olmocr, li2025monkeyocr}. Rasterized tables, visually embedded rules, and figures carrying information absent from the text remain difficult~\cite{ke2025large,dalvand2025indian}: rule-based methods rely on geometric heuristics, VLM-based ones add latency and instability on dense layouts~\cite{verbovskiy2025comparing}. Qwen-VL~\cite{bai2025qwen3, wang2024qwen2} and LLaVA~\cite{liu2023visual} extract structure well~\cite{dong2026doc, zhu2024mmdocbenchbenchmarkinglargevisionlanguage} but vary in hallucination, grounding, and throughput on image-heavy documents. These weaknesses compound in enterprise governance, where guidelines span every content type and parsing failures propagate into rule quality.

Prompting and tuning sharpen extraction itself: GoLLIE~\cite{sainz2024gollieannotationguidelinesimprove} shows guidelines in prompts improve zero-shot extraction, and UIE~\cite{lu2022unifiedstructuregenerationuniversal} improves robustness across event types, as as does discourse segmentation \cite{sediqin2025laces,11248014}  These systems, with parsers such as Docling~\cite{livathinos2025docling} and Donut~\cite{kim2022ocr}, target extraction alone, offering no validation, contradiction handling, or governed artifact generation; we therefore evaluate against a monolithic baseline sharing SAGE's structure (Section~\ref{sec:monolithic}).

Coordination is a separate concern. Multi-agent systems decompose tasks across specialized agents with distinct tools, memory, and protocols~\cite{react2023}, their effectiveness turning on shared state, intermediate results, and human oversight~\cite{autogen2023}. Existing frameworks pass unstructured messages; SAGE instead (i) coordinates stages through a shared versioned rule store with stable identifiers, analogous to materialized intermediate tables in a query pipeline~\cite{weiss1999, wang2024survey, herschel2017survey}; (ii) guarantees schema-validated outputs before the next stage consumes them; and (iii) escalates on a threshold-driven, persona-routed basis, with QM and PM workbenches receiving only items requiring their expertise.

\subsection{Human-in-the-Loop Workflows}
HITL systems improve annotation quality by combining model predictions with human verification~\cite{wu2021survey}, with effectiveness depending on selective routing, interface design, and escalation policy~\cite{weiss1999}. Existing approaches treat HITL as a flat review queue, routing uncertain outputs uniformly regardless of object type, severity, or downstream dependency. SAGE instead routes only rules and artifacts that fail structural or semantic thresholds, carry inferred source annotations, or are flagged as contradictions or duplicates, reusing zero-edit approvals as calibration signals to progressively reduce review load.

\section{System Architecture}
\label{sec:architecture}
\begin{figure*}[t]
\centering
\includegraphics[width=0.85\textwidth]{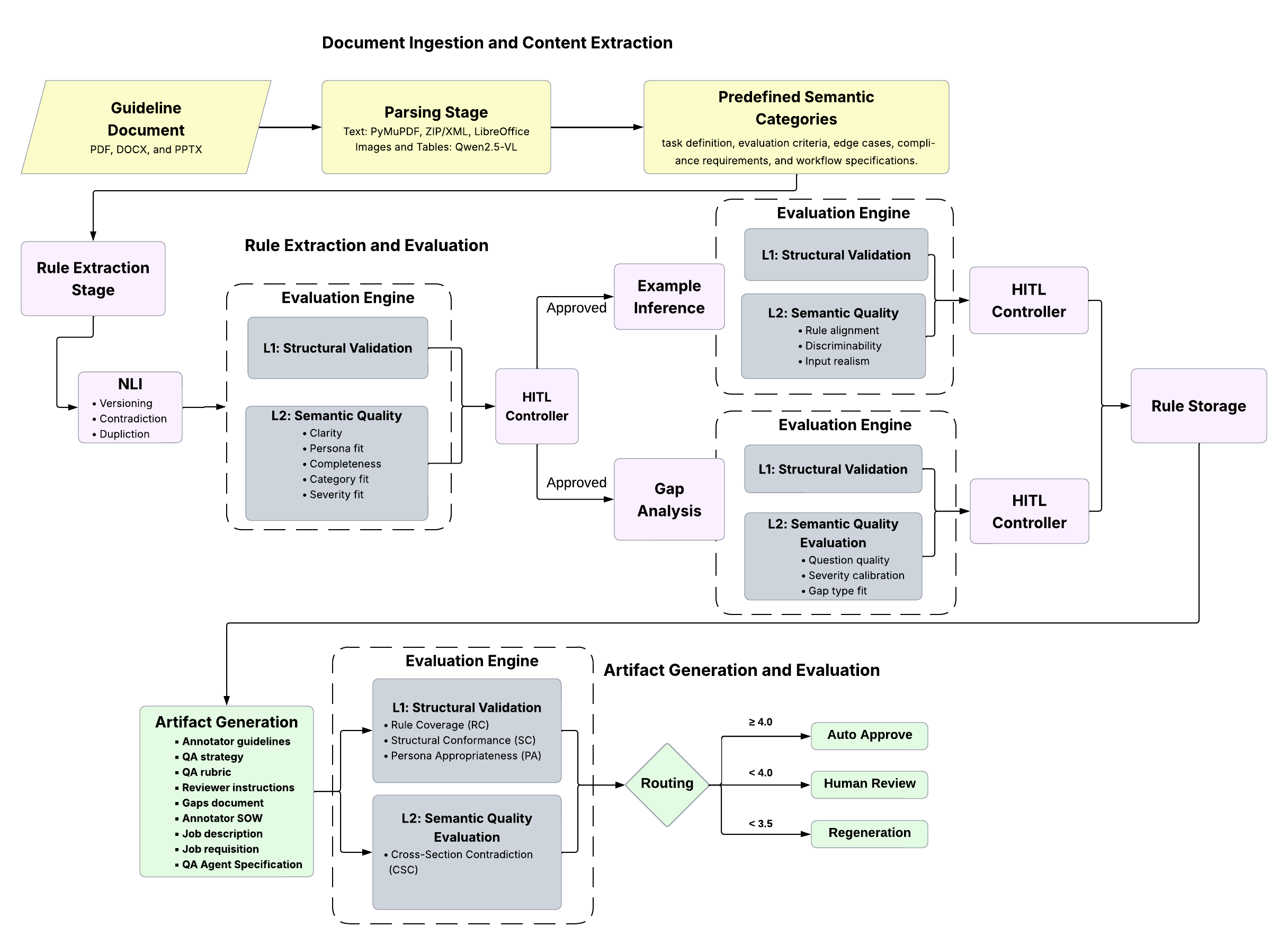}
\caption{Overview of the SAGE pipeline.}
\label{fig:pipeline_overview}
\end{figure*}

\textsc{SAGE} is architected around a central versioned rule store: a set of
schema-enforced tables, keyed by a stable \texttt{rule\_id}, that serves as the shared data layer for every stage. Rather than passing unstructured messages, stages read from and write to this store through schema-validated contracts (Pydantic models), so no downstream stage ever consumes structurally invalid data. This design yields three properties essential
for enterprise deployment: \textit{provenance}, as every artifact traces
back to its source rules and originating document; \textit{versioning}, as
rule updates across document revisions are reconciled rather than
reprocessed from scratch; and \textit{auditability}, as every HITL decision
is logged against a stable identifier for governance review. Concretely, the system comprises a Parsing stage, a Rule Extraction stage, a Consistency module, an Evaluation module, an HITL Controller, and an Artifact Generation stage. Documents flow through a structured pipeline of ingestion, rule extraction, parallel consistency checking and evaluation, threshold-driven HITL review, and artifact generation, with each stage strictly conditioned on finalized outputs from preceding stages to ensure dependency-aware refinement.
Figure~\ref{fig:pipeline_overview} presents an overview of the system.

\subsection{Document Ingestion and Content Extraction}
\label{sec:ingestion}
Enterprise guideline documents combine text, tables, and visual elements that require heterogeneous extraction strategies. The parsing stage addresses this through a two-stage architecture that separates deterministic text extraction from VLM processing, ensuring reproducibility while preserving full multimodal coverage across all document formats.

\textbf{Text extraction} is fully deterministic and does not involve any language model. PDFs are processed with PyMuPDF to extract text, layout, images, and structural metadata; DOCX files are parsed directly from their internal XML structures via ZIP-based access, enabling faithful reconstruction of hierarchical content; and PPTX files are normalized to PDF via LibreOffice conversion, then routed through the same PyMuPDF pipeline for format consistency.

\textbf{Visual extraction} follows a separate path. Images are deduplicated by MD5 hashing before VLM processing to remove redundancy, and tables are detected via layout analysis and processed with a verbatim-preserving prompt to maintain structural fidelity. All non-textual elements are handled by Qwen2.5-VL~\cite{wang2024qwen2}.
Extraction quality is assessed along two dimensions which are
coverage and quality score.
\textbf{Coverage} measures the fraction of source content units successfully extracted, reported separately for pages, figures, and tables. \textbf{Quality score (Q)} is captured using a composite scoring mechanism:
\begin{align}
    \label{aa}
    Q = 1 - (\text{garbage} + \text{mojibake} \nonumber \\
    + \text{repetition} + \text{silent\_skip})
\end{align}
where each term represents a normalized defect rate corresponding to extraction noise, encoding corruption, redundant outputs, and missing content respectively. The score $Q \in [0,1]$ provides an aggregate measure of extraction fidelity.

Extracted content is then segmented into predefined semantic categories including task definition, evaluation criteria, edge cases, compliance requirements, and workflow specifications. This segmentation serves as a routing layer for downstream rule extraction processing stage. All outputs are indexed by document and page and emitted as JSON containing the document text together with the descriptions generated for images and tables, which forms the input for the rule extraction stage.

\subsection{Rule Extraction}
\label{sec:rule_extraction}
The Rule Extraction module employs a two-stage inference pipeline powered by a Qwen-family vision-language model, applied here in text-only mode to the JSON produced by the parsing stage. The first stage performs open-domain extraction, identifying candidate rules from document content and associating each with a source span, rule type, and confidence score. The second stage normalizes each candidate into a fixed 26-field rule schema, ensuring consistency and downstream compatibility.
Rule type determines persona routing: the Quality Manager (QM) workbench receives \texttt{evaluation-criteria}, \texttt{edge-case}, and \texttt{qa-process} rules, and the Project Manager (PM) workbench receives \texttt{worker-requirements} and \texttt{delivery-schema} rules.

Once normalized, rules are first passed through the two-stage consistency module, which applies embedding-based similarity filtering followed by natural language inference (NLI)~\cite{maccartney2009natural} classification to perform version alignment, deduplication, and contradiction detection against the existing rule store. The Evaluation module then performs structural and semantic quality assessment on the remaining rules, producing pass, flag, or reject outcomes.
Rules flagged by either component are routed to HITL review. The HITL Controller jointly considers evaluation and consistency signals and routes rules requiring human judgment to the appropriate QM or PM workbench.

\subsubsection{Gap Analysis}
\label{sec:gap_analysis}
The system performs gap analysis against the approved rule set to identify missing, ambiguous, or underspecified aspects of the source guidelines. Each detected gap is represented as a structured \textit{GapObject}, containing a targeted clarification question that is surfaced in the HITL workbench for resolution by the appropriate stakeholder. Resolved gaps are converted into \textit{ClarificationRecord}s, which may generate additional \textit{RuleUnit}s when necessary. These updates are appended to the rule store, ensuring that downstream artifact generation operates over a complete and unambiguous specification.

\subsubsection{Example Inference}
\label{sec:example_inference}
For each approved rule, the system extracts examples from the source guideline when explicitly available. In cases where examples are not directly specified, contextually grounded examples may be inferred under strict adherence to the rule semantics.
All extracted and inferred examples are subjected to the same L1/L2 evaluation framework as rules prior to storage. Approved examples are then used to support annotator guidance and reviewer validation during ongoing task execution.

\subsection{Evaluation Engine}
\label{sec:evaluation_engine}
As ground truth annotations are unavailable for this corpus, all evaluation
metrics are computed using standard signals: cosine similarity for grounding
and hallucination rates, Pydantic-based schema validation for structural
compliance, and LLM-as-judge scoring for semantic quality. Outputs from a sample of 10--15 documents were manually inspected by a domain expert, confirming that scores were consistent with human judgment across extraction, rule quality, and artifact evaluation. All thresholds were selected empirically over the full corpus and fixed prior to all reported experiments, following standard practice for production system evaluation where held-out splitting would reduce the document diversity available for calibration.

\subsubsection{VLM Benchmarking}
\label{sec:vlm_benchmarking}
Prior to structured extraction, VLM outputs are evaluated across six dimensions to assess suitability for downstream processing. \textit{Evidence rate} is the fraction of \texttt{RuleUnit}s grounded in the source document, computed as $\max \text{cosine}(\text{rule, source sentences})$ using all-MiniLM-L6-v2, where a score $\geq 0.65$ indicates support; \textit{hallucination rate} is the fraction with no match, treating scores $< 0.40$ as hallucinated and $0.40$ to $0.65$ as ambiguous. \textit{Quality score} follows Eq.~\ref{aa}, and \textit{throughput} is the ratio of successfully processed images to total input images. \textit{Duplication rate} is the fraction of semantically equivalent \texttt{RuleUnit} pairs, identified by cosine $\geq 0.85$ followed by NLI $\geq 0.70$ (DeBERTa-v3-large-mnli), and \textit{category distribution} is the normalized Shannon entropy over rule categories.

\subsubsection{Rule Evaluation Framework}
\label{sec:rule_eval_framework}
All structured objects undergo a two-stage validation pipeline before being considered for HITL review.

\textbf{L1: Structural Validation.}
It performs deterministic schema verification using 26 Pydantic constraints covering required fields, type correctness, enumeration validity, and internal logical consistency. Only objects satisfying all constraints are passed to L2. Failed objects are either retried, corrected, or rejected depending on error severity. L1 is fully deterministic and does not involve any language model.

\textbf{L2: Semantic Evaluation.}
Objects passing L1 are evaluated using an LLM-as-judge framework with $K$ quality dimensions:
\begin{align}
   S(x) = \frac{1}{K} \sum_{k=1}^{K} s_k(x), \quad s_k \in \{1,2,3,4,5\}
\end{align}
where $s_k$ is the score assigned to the $k$-th dimension by the LLM judge, with $K = 5$ for \textit{RuleUnit} and $K = 3$ for both \textit{ExampleObject} and \textit{GapObject}. Routing is determined by the minimum dimension score rather than the average, ensuring that a single weak dimension cannot be masked by strong performance on others:

\begin{itemize}
    \item $\min_k s_k \geq 4$: auto-approve
    \item $s_k \in \{2,3\}$ for any $k$: route to HITL
    \item $s_k = 1$ for any $k$: reject
\end{itemize}

Objects with \texttt{rule\_source = inferred} are always routed to HITL regardless of score. Zero-edit HITL approvals are logged as calibration data for the L2 evaluator.

\subsubsection{Object-Specific Evaluation}
\label{sec:object_specific}
L1 and L2 dimensions are defined per object type as follows.

\textbf{RuleUnit.} L1 enforces 14 structural constraints, including required field presence, valid enumerations, and prohibition of instruction-source duplication. L2 then scores five dimensions: \textit{clarity}, whether instructions are unambiguous and imperative; \textit{persona fit}, alignment with the target \textit{applies\_to} audience; \textit{completeness}, coverage of all conditions and edge cases; \textit{category fit}, correctness of rule categorization; and \textit{severity fit}, proportionality between the assigned severity and the impact of a violation.

\textbf{ExampleObject.} L1 enforces 4 constraints, including valid rule linkage and separation of correct and incorrect outputs. L2 scores three dimensions: \textit{rule alignment}, whether the example directly and faithfully tests the linked rule; \textit{discriminability}, whether correct and incorrect outputs are clearly separated; and \textit{input realism}, whether the scenario is a plausible annotation case.

\textbf{GapObject.} L1 enforces 8 structural constraints, including valid gap types, required fields, and resolved rule references. L2 scores three dimensions: \textit{question quality}, whether the clarification question is specific enough to elicit a usable rule; \textit{severity calibration}, proportionality between gap risk and annotation impact; and \textit{gap type fit}, correctness of the categorical assignment.

\subsection{Human-in-the-Loop Review}
\label{sec:hitl_review}
SAGE enforces a staged dependency-aware workflow in which gap analysis and example inference are conditioned on the approved rule set, bounding the downstream review surface so that QM effort targets only gaps and examples grounded in approved rules. In Phase 1, the QM approves, rejects, or edits extracted \textit{RuleUnit}s in the QM workbench; the approved rule set forms the authoritative specification for all downstream processing and cannot be bypassed or modified implicitly in later phases. In Phase 2, \textit{GapObject}s (Section~\ref{sec:gap_analysis}) are reviewed by the QM, and approved gaps are resolved into \textit{ClarificationRecord}s, each spawning a new \texttt{RuleUnit} appended to the authoritative rule set. In Phase 3, \textit{ExampleObject}s (Section~\ref{sec:example_inference}) are reviewed conditioned on the finalized rule set and resolved gaps, routed to the QM or PM workbench via the \textit{owned\_by} tag of the linked \textit{RuleUnit}.

\subsection{Artifact Generation and Evaluation}
\label{sec:artifact_generation}
The artifact generation stage transforms approved
\textit{Rules (RuleUnits)}, resolved \textit{Gaps (GapObjects)}, and
evaluated \textit{Examples (ExampleObjects)} into nine
operational artifacts using Pydantic-constrained templates, generated per
persona from the approved rule store.

\begin{itemize}
\item \textbf{QM:} annotator guidelines, quality assessment (QA) strategy, QA rubric, reviewer instructions, gaps document, and QA agent specification \cite{kothari2026position}, consumed by annotators, reviewers, quality leads, and Learning \& Development
\item \textbf{PM:} annotator SOW, job description, and job requisition, consumed by Recruiting and crowd contributors
\end{itemize}

Each artifact undergoes a two-layer evaluation before release. L1 applies three automated metrics, each with a pass threshold of $\geq 3/5$. \textit{Rule coverage} (RC) is the percentage of approved \textit{RuleUnit}s reflected in artifact content, measured via cosine similarity between \textit{RuleUnit}s and artifact sections. \textit{Structural conformance} (SC) verifies that required sections are present, correctly ordered, and non-empty, using Python section matching. \textit{Persona appropriateness} (PA) checks that tone matches the target persona, combining Flesch readability score, grade level, and LLM judgment for clarity.
L2 applies a single cross-artifact metric: Cross-Section Contradiction (CSC),
in which an LLM reviews the full artifact for contradictions and inconsistent
\texttt{RuleUnit}s across sections (pass: $\geq 4/5$, evaluated using Qwen2.5-VL). The artifact routing threshold (ART) combines RC, SC, PA, and CSC with empirically tuned weights:
\begin{align}
\text{ART} = w_1 \cdot \text{RC} + w_2 \cdot \text{SC} +
w_3 \cdot \text{PA} + w_4 \cdot \text{CSC}
\end{align}
$\text{ART} \geq 4.0$ auto-approves the
artifact; $3.5 \leq \text{ART} < 4.0$ routes to human review;
$\text{ART} < 3.5$ blocks release and triggers regeneration.

\section{Results and Discussion}
\label{sec:results}

\subsection{Dataset}
\label{sec:dataset}
Our evaluation corpus consists of 120 enterprise guideline documents provided
by industrial clients under confidentiality agreements. All 120 documents are text, table, figure, and image-heavy enterprise guideline files in PDF, DOCX, or PPTX format, sharing a single parsing path. All documents are used as received without pre-processing, reflecting real production variability in structure and
complexity. Documents contain natural language rules, structured and
unstructured tables, and rich visual content including annotated images
with embedded text.

We define three complexity tiers by modality composition: Low (text-dominant, $\sim$20--30\,min), Moderate (text + images + tables, $\sim$30--65\,min), and High (fully multimodal, $\sim$65--100\,min). The corpus comprises 8 text-only documents, 84 with tables and figures, and 28 that are table- and image-heavy. Modality density is the primary driver of variance, as image-heavy and table-dense documents incur additional VLM passes and higher HITL escalation.

\subsection{VLM Selection and Parsing Evaluation}
\label{sec:vlm_selection}
We evaluate Qwen2.5-VL-32B, Qwen3-32B, and LLaVA-13B on the full 120-document corpus under identical settings, using the metrics of Section~\ref{sec:vlm_benchmarking}. To avoid circular dependency with the system parser (PyMuPDF), extraction baselines are computed independently with pdfminer, pypdf, and pdfplumber, aggregating page counts by median, image counts by majority vote, and text by union for maximum coverage consistency.

Prioritising grounding and consistency over raw generation quality, we select Qwen2.5-VL-32B for all extraction stages: Table~\ref{tab:vlm_main} shows it leading on evidence rate, hallucination, throughput, and duplication. Qwen3-32B scores higher on quality and category distribution, reflecting improved reasoning behaviour, but its higher hallucination and duplication would push more rules into consistency filtering and human review; LLaVA-13B underperforms across all grounding and consistency dimensions. These are raw extraction rates measured before governance; the 3.2\% reported in Section~\ref{sec:rule_eval} is post-governance. Docling~\cite{livathinos2025docling} was also evaluated as an alternative parser, but its output proved insufficiently structured for fine-grained image-detail extraction, motivating the VLM-based path.

\begin{table}[htbp]
\caption{VLM benchmark on the 120 documents. Quality score is Eq.~\ref{aa}
applied to raw VLM output; throughput is images processed out of 355.}
\label{tab:vlm_main}
\centering
\small
\setlength{\tabcolsep}{4pt}
\renewcommand{\arraystretch}{1.1}
\begin{tabular}{@{}lrrr@{}}
\toprule
Metric & \textbf{Qwen2.5} & Qwen3 & LLaVA \\
       & \textbf{-VL-32B} & -32B  & -13B \\
\midrule
Evidence rate (\%)   & \textbf{77.8} & 76.7 & 64.1 \\
Hallucination (\%)   & \textbf{20.2} & 22.7 & 35.9 \\
Quality score (\%)   & 64.9 & \textbf{73.2} & 63.6 \\
Throughput (img/355) & \textbf{232}  & 184  & 159 \\
Duplication (\%)     & \textbf{14.6} & 17.9 & 42.0 \\
Category dist.\ (\%) & 51.5 & \textbf{54.5} & 42.9 \\
\bottomrule
\end{tabular}
\end{table}

Table~\ref{tab:content_summary} reports 99.2\% page coverage, 97.1\% figure recall, 88.3\% table recall, and 96\% (115/120) document success. Near-zero defect ratios leave the Eq.~\ref{aa} score at 1.00 on all successful documents; the five failures are VLM timeouts on image-heavy documents and poorly structured tables exceeding the processing budget.

\begin{table}[htbp]
\caption{Content extraction evaluation: 120 documents. Overall score is
Eq.~\ref{aa} applied to the full parsing pipeline.}
\label{tab:content_summary}
\centering
\small
\setlength{\tabcolsep}{3pt}
\begin{tabular}{llc}
\toprule
Category & Metric & Value \\
\midrule
Coverage
& Page coverage     & 99.2\% \\
& Figure recall     & 97.1\% \\
& Table recall      & 88.3\% \\
\midrule
Quality score
& Overall score     & 1.00   \\
& Garbage ratio     & 0.00\% \\
& Mojibake ratio    & 0.00\% \\
& Repetition ratio  & 0.01\% \\
& Silent skip ratio & 0.00\% \\
\midrule
Success rate
& Document success  & 96\% (115/120) \\
\bottomrule
\end{tabular}
\end{table}

\subsection{Rule Extraction Evaluation}
\label{sec:rule_eval}
Rule extraction is evaluated on the 115 documents that completed content extraction, yielding 3,896 \texttt{RuleUnit}s, following the framework in Section~\ref{sec:evaluation_engine}. Table~\ref{tab:rule_summary} reports an evidence rate of 84.8\%, coverage of 82.6\%, and a hallucination rate of 3.2\%. L1 passes 99.1\% of units; the 0.9\% flagged for ambiguity are structurally valid but lack sufficient semantic precision for direct execution. At L2, 71.4\% of \texttt{RuleUnit}s are auto-approved, 28.6\% are routed to HITL review, and 0.0\% are rejected: rejection occurs only when a rule fails structural validation or receives the minimum score on an L2 dimension, and by design uncertain cases are deferred to human review rather than discarded. The consistency module identifies gaps in 26.7\% of units, duplications in 3.0\%, and contradictions in 2.9\%.

To evaluate whether auto-approval reflects quality rather than leniency, we validate the L2 judge against a blind evaluation set of 300 rule-level annotations, each independently labeled by expert annotators without access to the judge outputs. The judge achieves a precision of 0.941, recall of 0.974, and F1 of 0.957, with a raw agreement of 93.33\% and a Cohen's $\kappa$ of 0.813. The results suggest alignment between judge decisions and human annotations.

To quantify each governance layer, Table~\ref{tab:ablation} ablates over the same 3,896 extracted RuleUnits. Without governance, every unit is auto-approved unverified, including 117 latent duplicates. Adding L1/L2 evaluation with HITL routing sends 28.6\% of units to human review and auto-approves the rest, while retention is unchanged: the system gates and routes rather than deletes. The consistency module then discards the 117 duplicates, 84 previously auto-approved and 33 already queued for review, reducing retained rules to 3,779 and the queue to 1,081. Within that queue it flags 113 contradictions for resolution, and it surfaces 1,040 gaps that would otherwise go undetected. Each layer thus contributes a distinct, measurable effect.

\begin{table}[htbp]
\caption{Governance-layer ablation (115 documents, 3,896 RuleUnits).
Columns are cumulative.}
\label{tab:ablation}
\centering
\small
\setlength{\tabcolsep}{4pt}
\renewcommand{\arraystretch}{1.15}
\begin{tabular}{@{}lrrr@{}}
\toprule
Metric & None & +L1/L2 & +Cons. \\
\midrule
Rules retained          & 3,896 & 3,896 & 3,779 \\
Auto-approved           & 3,896 & 2,782 & 2,698 \\
Routed to HITL          & 0     & 1,114 & 1,081 \\
Duplicates present      & 117   & 117   & 0 \\
Contradictions surfaced & 0     & 0     & 113 \\
Gaps surfaced           & 0     & 0     & 1,040 \\
\bottomrule
\end{tabular}
\end{table}

\begin{table}[htbp]
\caption{Rule extraction: 115 documents, 3,896 RuleUnits.}
\label{tab:rule_summary}
\centering
\small
\begin{tabular}{llc}
\toprule
Category & Metric & Value \\
\midrule
Quality
& Evidence rate & 84.8\% \\
& Coverage      & 82.6\% \\
& Hallucination & 3.2\% \\
\midrule
L1
& Pass rate     & 99.1\% \\
& Ambiguity     & 0.9\%  \\
\midrule
L2
& Auto approved & 71.4\% \\
& Human review  & 28.6\% \\
& Rejected      & 0.0\%  \\
\midrule
Consistency
& Gaps          & 26.7\% \\
& Duplications  & 3.0\%  \\
& Contradictions& 2.9\%  \\
\bottomrule
\end{tabular}
\end{table}

\subsection{Artifact Generation Evaluation}
\label{sec:artifact_eval}
We evaluate artifact generation over the approved rule set from the same 115 documents, producing 812 artifacts spanning nine artifact types. Each type is generated only where the approved rule set contains the rule types it draws on, so the number of artifacts per document varies with the composition of its guidelines. Artifacts are generated using Claude Sonnet 4.6 and evaluated using Qwen2.5-VL against a fixed rubric-based
prompting strategy following the metrics defined in Section~\ref{sec:artifact_generation}, ensuring
generation and evaluation are performed by independent model families
to avoid self-evaluation bias.
Table~\ref{tab:artifact_summary} shows strong artifact quality across
dimensions. Rule coverage (fraction of \texttt{RuleUnit}s reflected in
artifact) reaches 93.7\%, indicating that artifacts faithfully instantiate
the approved rules, while cross-section contradiction (logical consistency
across sections) at 96.3\% confirms high logical consistency. Structural
conformance at 82.4\% and persona appropriateness (tone-audience
alignment) at 81.0\% show that generated artifacts follow the required
section structure and align well with their target audience. Overall, 54.2\% of artifacts are
auto-approved and only 3.0\% are rejected, with the remaining 42.8\% routed
to human review, reflecting a substantial reduction in review burden.

Coverage is computed by nearest-section cosine similarity (all-MiniLM-L6-v2), with no LLM judgment. We inspected the full below-threshold set and found no true omissions: those rules are either realized in a different artifact type where more naturally expressed (e.g. output-schema constraints in schema-focused artifacts) or out of scope for the artifact type by design. Near-threshold cases, which we do not audit exhaustively, are largely paraphrastic. We therefore treat rule coverage as a conservative lower bound rather than a completeness guarantee, and escalate all artifacts below the L1 threshold for review.

\begin{table}[htbp]
\caption{Artifact evaluation: 115 documents, 812 artifacts.}
\label{tab:artifact_summary}
\centering
\small
\setlength{\tabcolsep}{3pt}
\renewcommand{\arraystretch}{1.0}
\begin{tabular}{llc}
\toprule
Category & Metric & Value \\
\midrule
HITL routing
& Auto approved (green) & 54.2\% \\
& Human review (amber)  & 42.8\% \\
& Rejected (red)        & 3.0\% \\
\midrule
L1 metrics
& Rule coverage & 93.7\% \\
& Structural conformance & 82.4\% \\
& Persona appropriateness & 81.0\% \\
\midrule
L2 metrics
& Cross-section contradiction & 96.3\% \\
\bottomrule
\end{tabular}
\vspace{-10pt}
\end{table}

\subsection{Comparison Against a Monolithic Baseline}
\label{sec:monolithic}
To isolate the contribution of the governance pipeline, we compare SAGE against a monolithic one-pass baseline that produces rules and artifacts directly from the parsed document, without the rule store, L1/L2 validation, consistency module, or HITL routing. Both configurations use the same parser and the same models at each stage, Qwen2.5-VL-32B for rule extraction and Claude Sonnet 4.6 for artifact generation, and are scored with an identical harness on the same 115 documents; this baseline is therefore distinct
from the VLM benchmark in Table~\ref{tab:vlm_main}, which measures raw extraction for
model selection. As Table~\ref{tab:baseline} shows, removing governance degrades every
metric: hallucination rises from 3.2\% to 15.7\%, duplication from 3.0\% to 10.3\%, and
artifact rule coverage falls from 93.7\% to 62.8\%, confirming that output quality is
driven by the governance pipeline rather than the base model alone.

\begin{table}[htbp]
\caption{Monolithic one-pass baseline vs.\ SAGE. }
\label{tab:baseline}
\centering
\small
\setlength{\tabcolsep}{4pt}
\renewcommand{\arraystretch}{1.15}
\begin{tabular}{@{}lcc@{}}
\toprule
Metric & Baseline & SAGE \\
\midrule
Hallucination               & 15.7\% & \textbf{3.2\%} \\
Duplication                 & 10.3\% & \textbf{3.0\%} \\
Contradiction               & 7.8\%  & \textbf{2.9\%} \\
L1 pass rate                & 93.2\% & \textbf{99.1\%} \\
Rule coverage               & 62.8\% & \textbf{93.7\%} \\
Persona appropriateness     & 37.7\% & \textbf{81.0\%} \\
Structural conformance      & 79.4\% & \textbf{82.4\%} \\
Cross-section contradiction & 94.2\% & \textbf{96.3\%} \\
\bottomrule
\end{tabular}
\end{table}

\subsection{End-to-End Effort}
\label{sec:effort}
The quality and project managers who currently produce these artifacts by hand estimate 2--3 working days per document, or 1,423 minutes for a single complex, image- and table-heavy one. SAGE completes the same document in 70--85 minutes. Appendix~\ref{app:effort} reports the per-stage breakdown.

\section{Conclusion}
\label{sec:conclusion}
We proposed SAGE, a governed multi-stage LLM pipeline that coordinates parsing, extraction, evaluation, and artifact generation through a shared versioned rule store, converting heterogeneous enterprise guideline documents into validated, structured operational artifacts with full provenance. SAGE keeps human review selective through threshold-based escalation, and zero-edit approvals are fed back to sharpen the evaluation judge over time. Experiments on real-world enterprise documents confirm its effectiveness across content extraction, grounding quality, structural validation, and consistency enforcement, providing a scalable basis for governed enterprise document understanding. Next steps include improving table detection on degraded layouts, learning persona adaptation from approved artifacts, and extending SAGE to multilingual settings and to legal, clinical, and regulatory domains.


\section*{Limitations}

SAGE demonstrates strong performance across content
extraction, rule generation, consistency validation, and artifact generation
on real-world enterprise documents. VLM extraction stability decreases on
low-quality scans and borderless or merged-cell tables, a known challenge
across current vision-language systems. Persona appropriateness and rule
coverage remain the most challenging artifact dimensions, reflecting the
inherent difficulty of adapting technical rules to non-expert audiences.
The calibration mechanism relies on accumulating zero-edit HITL approvals
over deployment cycles, and scoring stability in early cycles remains
limited. The current evaluation also covers English enterprise guidelines only.


\bibliography{reference}

\clearpage
\appendix

\setcounter{table}{0}
\renewcommand{\thetable}{A\arabic{table}}

\section{Appendix}

\subsection{Per-Stage Effort Breakdown}
\label{app:effort}

Table~\ref{tab:effort_stages} maps each manual work step to the SAGE stages that replace it, for the document discussed in Section~\ref{sec:effort}. Reading and understanding the guideline corresponds to ingestion and content extraction (Section~\ref{sec:ingestion}); inferring and structuring rules to rule extraction (Section~\ref{sec:rule_extraction}), L1/L2 evaluation (Section~\ref{sec:evaluation_engine}), and Phase~1 rule review (Section~\ref{sec:hitl_review}); resolving ambiguities, conflicts, and gaps to the consistency check (Section~\ref{sec:rule_extraction}), gap analysis (Section~\ref{sec:gap_analysis}), and Phase~2 gap review (Section~\ref{sec:hitl_review}); and drafting and validating artifacts to artifact generation and evaluation (Section~\ref{sec:artifact_generation}) and Phase~3 artifact review (Section~\ref{sec:hitl_review}).

Of the 70--85 minutes, roughly 47--54 are machine time and 23--31 are selective human review, so human effort concentrates in rule, gap, and artifact review rather than spreading uniformly across the pipeline. Across the evaluated documents, total effort ranges from roughly 20 to 100 minutes with the same machine-plus-review split. As the manual baseline is an expert estimate rather than a controlled measurement, the comparison indicates an order-of-magnitude reduction rather than an exact head-to-head.

\begin{table}[htbp]
\caption{Per-stage effort for a single complex (image- and table-heavy) document, separating machine from HITL time, as ranges across repeated runs.}
\label{tab:effort_stages}
\centering
\setlength{\tabcolsep}{1pt}
\renewcommand{\arraystretch}{1.15}
\resizebox{\columnwidth}{!}{%
\begin{tabular}{@{}lcccc@{}}
\toprule
Work step & Manual & Machine & HITL & Total \\
          & (min) & (min)   & (min) & (min) \\
\midrule
Read \& understand           & 240 & $\sim$20.7--23.7 & 0              & $\sim$20.7--23.7 \\
Infer \& structure rules     & 433 & $\sim$8.0--9.2   & $\sim$7.0--8.9  & $\sim$15.0--18.1 \\
Resolve ambiguities \& gaps  & 290 & $\sim$4.4--5.0   & $\sim$8.0--12.0 & $\sim$12.4--17.0 \\
Draft \& validate artifacts  & 460 & $\sim$14.0--16.0 & $\sim$8.0--10.1 & $\sim$22.0--26.1 \\
\midrule
\textbf{Total} & \textbf{1,423} & \textbf{$\sim$47.1--54.0} & \textbf{$\sim$23.0--31.0} & \textbf{$\sim$70.1--85.0} \\
\bottomrule
\end{tabular}}
\end{table}

\end{document}